\documentclass[runningheads]{llncs}

\usepackage[acronym]{glossaries}
\makeglossaries
\newacronym{auc}{AUC}{Area Under the Curve}
\newacronym{ale}{ALE}{Arcade Learning Environment}
\newacronym{cgp}{CGP}{Cartesian GP}
\newacronym{dl}{DL}{Deep Learning}
\newacronym{dqn}{DQN}{Deep Q-Network}
\newacronym{drl}{DRL}{Deep RL}
\newacronym{gp}{GP}{Genetic Programming}
\newacronym{lgp}{LGP}{Linear GP}
\newacronym{maple}{MAPLE}{Multi-Action Programs through Linear Evolution}
\newacronym{matpg}{MATPG}{Multi-Action TPG}
\newacronym{mac}{MAC}{multiply–accumulate}
\newacronym{ml}{ML}{Machine Learning}
\newacronym{mujoco}{MuJoCo}{Multi-Joint dynamics with Contact}
\newacronym{mtrl}{MTRL}{Multi-Task RL}
\newacronym{ppo}{PPO}{Proximal Policy Optimization}
\newacronym{rl}{RL}{Reinforcement Learning}
\newacronym{sac}{SAC}{Soft Actor-Critic}
\newacronym{sbb}{SBB}{Symbiotic Bid-Based}
\newacronym{td3}{TD3}{Twin Delayed Deep Deterministic}
\newacronym{tgp}{TGP}{Tree-Based GP}
\newacronym{tpg}{TPG}{Tangled Program Graph}

\usepackage{wrapfig}
\usepackage{graphicx}
\usepackage{algpseudocode}
\usepackage{algorithm}
\usepackage{amsfonts}
\usepackage{subcaption}
\usepackage{booktabs}
\usepackage{float}
\usepackage{caption}
\usepackage{stfloats}
\usepackage{amsmath}
\usepackage{tabularx}
\usepackage{comment}
\usepackage{hyperref}
\usepackage{orcidlink}
\usepackage{verbatim}

\usepackage[T1]{fontenc}
\usepackage{xcolor}
\newcommand{\kd}[1]{\textcolor{green}{[K]}}
\newcommand{\md}[1]{\textcolor{red}{[M]}}
\newcommand{\nb}[1]{\textcolor{blue}{[N]}}
\newcommand{\tm}[1]{\textcolor{yellow}{[T]}}
\newcommand{\pa}[1]{\textcolor{purple}{[P]}}
\newcommand{\qv}[1]{\textcolor{orange}{[Q]}}

\usepackage[normalem]{ulem}

\begin{document}
\title{Multi-Action Tangled Program Graphs for Multi-Task Reinforcement Learning with Continuous Control}

\titlerunning{Multi-Action TPG for continuous Multi-Task RL}
%
%
%
%

\author{Quentin Vacher\inst{1}\orcidlink{0009-0001-9568-7196} \and
Nicolas Beuve\inst{1}\orcidlink{0000-0002-1371-4016} \and
Micka\"el Dardaillon\inst{1}\orcidlink{0000-0001-6862-2090} \and 
Karol Desnos\inst{1}\orcidlink{0000-0003-1527-9668}}
\authorrunning{Q. Vacher et al.}
%
\institute{Univ Rennes, INSA Rennes, CNRS, IETR – UMR 6164, F-35000 Rennes, France
\email{first.last@insa-rennes.fr}}

\maketitle              
\begin{abstract}

Over the past few decades, machine learning has been widely used to learn complex tasks. 
\gls{rl}, inspired by human behavior, is a great example, as it involves developing specific behaviours for specific tasks. 
To further challenge algorithms, \gls{mtrl} environments have been introduced, requiring a single model to learn multiple behaviors.

The \gls{tpg} algorithm is a \gls{gp} algorithm designed for discrete \gls{mtrl} environments. 
Recently, the \acrshort{maple} algorithm has been proposed, as another \gls{gp} algorithm that achieves high results in single task continuous \gls{rl} environments.
A variation of the \gls{tpg} is proposed alongside \acrshort{maple}, named \gls{matpg} that aggregates \acrshort{maple} agents, and creates a control flow to activate them.
Initially tested on single task \gls{rl} environments only, \gls{matpg} achieved similar results to \acrshort{maple}.

In this work, we present a new benchmark based on the MuJoCo \textit{Half Cheetah} from Gymnasium. 
This benchmark features five distinct obstacles that are randomly positioned in front of the agent, each of which demands a unique behavior. 
This benchmark serves as a use case for \gls{matpg}, to prove its ability as a \gls{gp} solution for continuous \gls{mtrl} environments.
Our experiments demonstrate its superiority in this multi-task use case when combined with lexicase selection.
Furthermore, we examine the interpretability of the evolved graph, revealing that the decision flow of the model is fully interpretable.

\keywords{Multi-task Learning  \and Reinforcement Learning \and Tangled Program Graphs \and Continuous Control}
\end{abstract}
\section{Introduction}
Intelligence can be measured by the ability to solve diverse tasks.
Artificial intelligence has explored learning multiple tasks with single models, particularly in \gls{rl}, where agents reason over long-term consequences, solving the temporal credit assignment.
In \gls{mtrl}, this challenge extends across tasks, requiring deep environmental understanding.

Several \gls{mtrl} benchmarks capture these challenges~\cite{mujocoMTBench}.
In continuous control, many use the \gls{mujoco} suite~\cite{mujoco} from Gymnasium~\cite{gymnasium}, by modifying body parts or controlling multiple agents~\cite{mujocoMT}.
Most solutions employ deep learning, specifically \gls{drl}~\cite{td3,sac,dqn,ppo}, using transfer learning to share knowledge.
While \gls{drl} performs well, its architectures are complex and opaque, producing large, uninterpretable models~\cite{de2025evolution,llmario,explainableRL}.

Alternative machine learning architectures, such as \gls{gp}, solve \gls{rl} tasks while producing compact and interpretable solutions.  
Algorithms like \gls{lgp}~\cite{lgp}, \gls{tgp}~\cite{gpKoza}, and \gls{cgp}~\cite{cgp} handle continuous control, while \gls{tpg}~\cite{tpg} excels in discrete \gls{mtrl}~\cite{tpgmt}.
In~\cite{maple}, \gls{matpg} was proposed as a continuous adaptation of \gls{tpg}.
However, it does not outperform its simplified version, \gls{maple}, on single continuous control tasks.

We evaluate the effectiveness of \gls{matpg} in continuous-control \gls{mtrl} settings, comparing it against its simplified version, \gls{maple}, under both tournament and lexicase selection~\cite{tournament,lexicase}.
Tournament selection is the default method for \gls{maple} and \gls{matpg} in~\cite{maple}, while lexicase selection, originally designed for multi-task learning in classification and regression, encourages task-specific elites.
Our experiments demonstrate that both \gls{matpg} and \gls{maple} with lexicase selection outperform tournament selection in producing agents capable of solving individual tasks.
Furthermore, \gls{matpg} proves highly effective in continuous-control \gls{mtrl} environments, achieving significantly better performance than the other three configurations when evaluated on all tasks simultaneously with lexicase selection.

\begin{wrapfigure}{R}{0.5\textwidth}
\vspace{-10pt}
  \centering
  \includegraphics[width=0.48\textwidth]{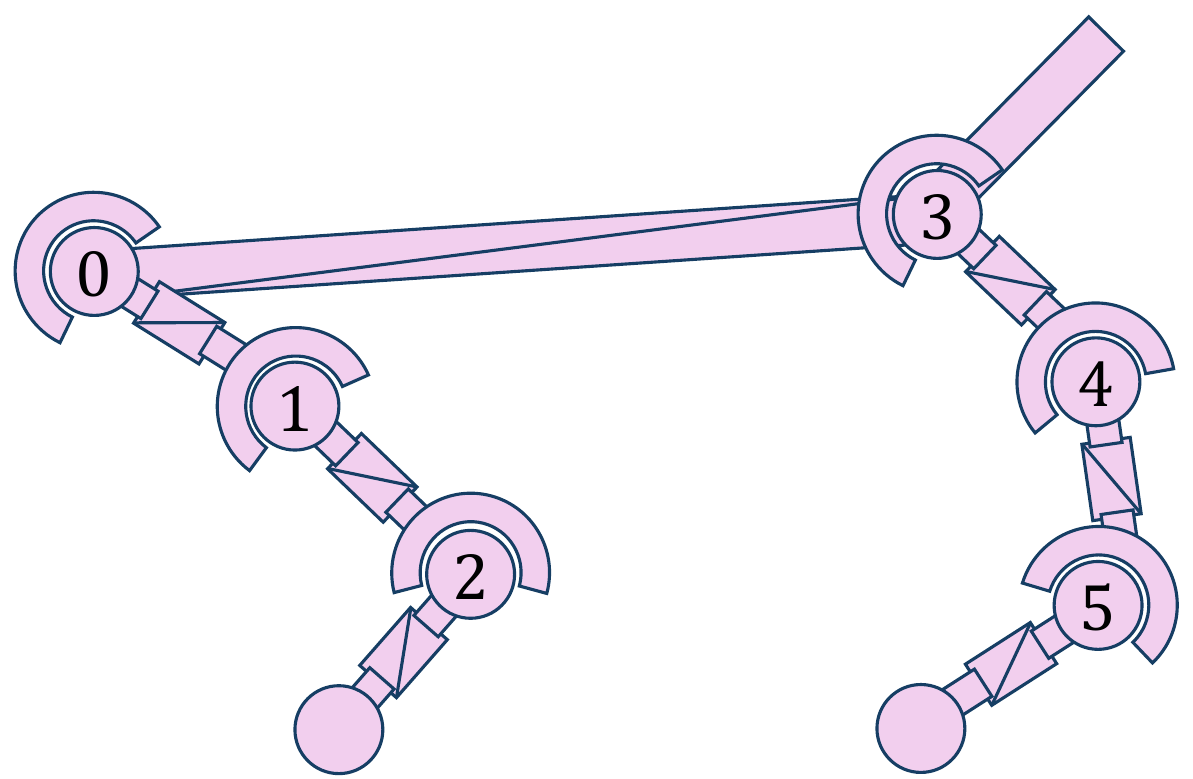} 
  \caption{Scheme of the \textit{Half Cheetah} environments from the \gls{mujoco} suite of Gymnasium.}
  \label{fig:half_cheetah}
\vspace{-6pt}
\end{wrapfigure}

To assess these algorithms, we designed a \gls{mtrl} benchmark based on the \gls{mujoco} \textit{Half Cheetah} environment, shown in Figure~\ref{fig:half_cheetah}, extended with five independent obstacles that an agent must overcome.
This benchmark is modular, meaning that learning one task does not enable the agent to overcome another without explicit learning.
It has been verified that the five obstacles can be learned independently by a single agent, making it a robust testbed for evaluating \gls{matpg} and \gls{maple}.
All the code and results used in this study are available on~\cite{QVACHER_EUROGP_2026}.

Section~\ref{sec:related_work} reviews prior work on \gls{mtrl} environments and \gls{gp} algorithms applied to \gls{rl}.  
Section~\ref{sec:background} provides a detailed description of the \gls{matpg} algorithm.  
Our proposed benchmark, based on a customized \textit{Half Cheetah}, is presented in Section~\ref{sec:use_case}, followed by the experimental setup in Section~\ref{sec:experimental_setup}, including the selection algorithms and parameter settings.  
Results are reported in Section~\ref{sec:experiments}, and Section~\ref{sec:interpretability} explores the policy of the best \gls{matpg} agent, highlighting its interpretability.  
Finally, Section~\ref{sec:conclusion} concludes the paper and discusses the findings.

\section{Related Work}
\label{sec:related_work}

To motivate the need for a new \gls{mtrl} benchmark, we provide a review of existing benchmarks in \gls{rl} in Section~\ref{sec:related_work_mtrl}. 
In this work, we challenge the proposed benchmark using the \gls{matpg} algorithm, which allows explicit reasoning over multiple tasks.
A complementary review of \gls{gp} algorithms applied to \gls{rl} is presented in Section~\ref{sec:related_work_gprl}, leading up to the development of \gls{matpg}.

\subsection{Multi-task Reinforcement Learning for Continuous Control}
\label{sec:related_work_mtrl}

Recent advances in \acrfull{rl} span diverse domains including robotics, games, autonomous systems, and NLP~\cite{rlsurvey}. This has driven development of \gls{drl} algorithms for continuous control like \gls{sac}~\cite{sac}, \gls{ppo}~\cite{ppo}, and \gls{td3}~\cite{td3}. 
Continuous control benchmarks, primarily in Gymnasium~\cite{gymnasium}, often use the \gls{mujoco} physics engine~\cite{mujoco}.
This benchmark comprises different robotic tasks that aim to reproduce real-world physics simulations.
Thus, the \gls{mujoco} benchmark is highly effective for scalable, realistic, single-task continuous control with \gls{rl}.

To evaluate multi-task capabilities, benchmarks requiring simultaneous task learning have emerged, leading to specialized \gls{mtrl} algorithms. 
IMPALA~\cite{impala} learns 57 \gls{ale} games with discrete actions, while continuous control \gls{mtrl} uses \gls{mujoco}, modifying \textit{Half Cheetah} morphology~\cite{mujocoMT,mujocoMTBench} or learning six tasks simultaneously with single-task performance.

However, \gls{drl} suffers from interpretability issues due to its \gls{dl} architecture~\cite{explainableRL}. For instance, a standard \gls{mtrl} network with two 400-unit layers exceeds 100k \gls{mac} operations, complicating understanding and increasing complexity~\cite{mujocoMT}. Alternative architectures like \gls{gp} address this by evolving compact, interpretable policies.

\subsection{Genetic Programming for Reinforcement Learning}
\label{sec:related_work_gprl}

\begin{figure}[t]
  \begin{center}
    \includegraphics[width=0.8\textwidth]{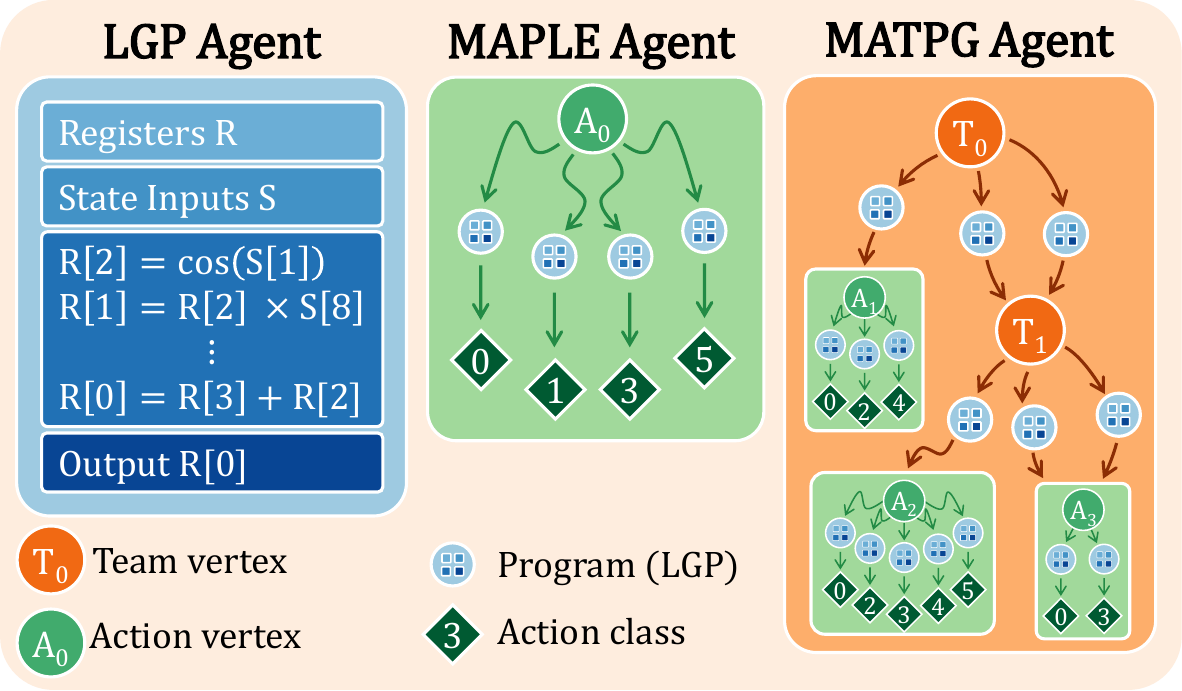}
  \end{center}
  \caption{Examples of \gls{lgp}, \gls{maple} and \gls{matpg} agents for an environment with six continuous actions. An action vertex uses all the output values of its programs to take the continuous actions, while a team vertex only follow the path of the program outputting the highest value.}
  \label{fig:matpg}
\end{figure}

As in many other areas, \acrfull{gp}~\cite{gpKoza} has been used to learn \gls{rl} tasks.
\gls{gp} can significantly improve the interpretability~\cite{de2025evolution} and reduce the computational complexity of a solution by orders of magnitude~\cite{vacherECTA} compared to \gls{drl}. 
In~\cite{lgpcartpole}, the \gls{lgp}~\cite{lgp} and \gls{tgp}~\cite{gpKoza} representations are used to learn different discrete tasks in the Gymansium library~\cite{gymnasium} such as the \textit{CartPole} or the \textit{Lunar Lander}, as well as some continuous tasks, such as \textit{Hopper} or the \textit{Bidepal Walker}.
In~\cite{nadizarMujoco}, \gls{lgp} ant \gls{cgp}~\cite{cgp} representations are trained independently on eight \gls{mujoco} environments.
Compared to state-of-the-art \gls{drl} algorithms, the \gls{gp} solutions achieve equivalent or better results on the tasks with three or fewer actions.
However, when the number of actions increases, \gls{drl} solutions become superior.

For discrete \gls{mtrl}, \gls{tpg} evolved a graph of \gls{lgp} programs, initially achieving \gls{dqn}-level performance on \gls{ale} with single Atari game, and later learning multiple Atari games with one model~\cite{tpg,tpgmt}. 
It was subsequently extended to continuous \gls{rl} by replacing discrete actions with an \gls{lgp} program outputting continuous values~\cite{heywoodwalker}.
To scale to many actions, two related approaches are proposed in~\cite{maple}: \gls{maple} and \gls{matpg} employing multiple \gls{lgp} programs, one per action. 
\gls{maple} is a flat multi-program policy, while \gls{matpg} integrates this into the hierarchical structure of \gls{tpg}.

On \gls{mujoco} benchmarks, \gls{maple} generally outperformed the continuous \gls{tpg} and matched \gls{matpg}, but with a smaller model size. 
The authors note that standard \gls{mujoco} tasks do not leverage key strength of \gls{matpg}: multi-task learning~\cite{maple}. 
This paper therefore investigates a dedicated \gls{mtrl} use case for \gls{matpg}.
Before doing so, the \gls{matpg} algorithm is presented with more details in the next section.

\begin{wrapfigure}{R}{0.5\textwidth}
\vspace{-20pt}
  \centering
  \includegraphics[width=0.48\textwidth]{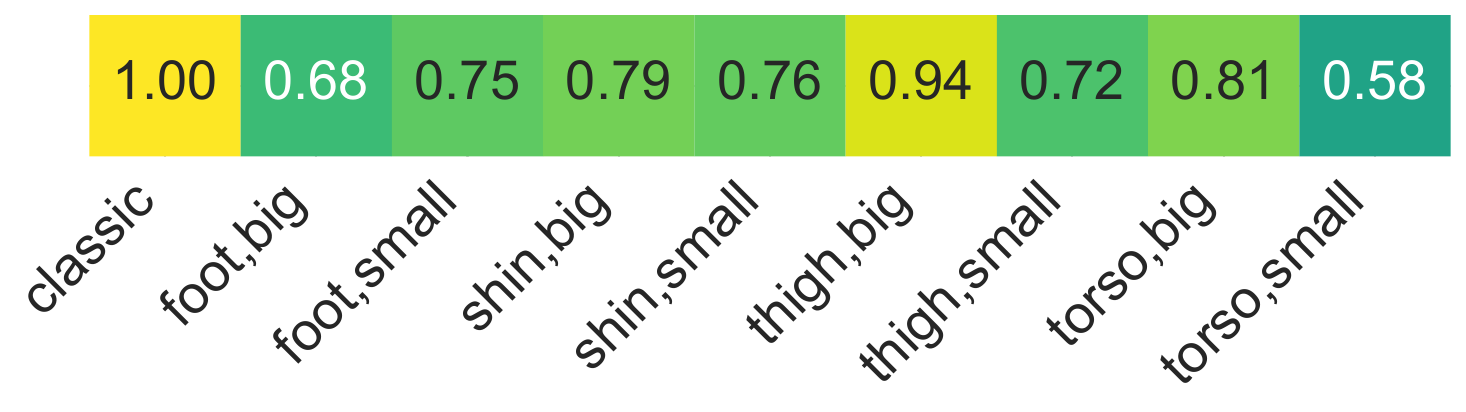} 
  \caption{Scores achieved by maple agents trained on classic \textit{Half Cheetah} but tested with modified body size.
           Scores are normalized by the unmodified score of 3622.}
  \label{fig:singleBodyPart}
\vspace{-12pt}
\end{wrapfigure}

\section{Background: Multi-Action Tangled Program Graphs}
\label{sec:background}

\subsubsection{Semantics of \gls{matpg}.}
\gls{matpg}~\cite{tpg,maple} consists of three main entities forming a direct graph: teams, actions, and programs.
Note that \gls{maple} is a specific case of \gls{matpg} without teams.
Figure~\ref{fig:matpg} shows an example of \gls{lgp}, \gls{maple} and \gls{matpg} agent.
Each team and action vertex contains multiple edges with programs corresponding to independent \gls{lgp}. 
Team programs point to another vertex, while action programs point to an action class.
Execution begins at a designated team, called root team.
This team activates all its programs with the same state inputs from the observation of the environment.
The destination with the highest program output is followed.
If it points to another team, the process repeats.
If it points to an action vertex, that vertex activates its programs.
Each program outputs a continuous value defining the final action for its class.
Missing action classes use a default action.
The actions taken are then sent to the environments.

\subsubsection{Mutation process.}
The graph structure evolves dynamically.
Each generation creates a fixed number of roots by duplicating and mutating the survivors of natural selection process.
Mutation can add or remove edges, or mutate existing ones.
When mutating an edge, either its program or destination can be mutated.
If the edge led to an action vertex, that vertex can be mutated instead of the program.

An action vertex is a \gls{maple} agent which is mutated in a similar fashion to team vertex, with the additional constraint that than any action class can appear only once in the action class edges.

\subsubsection{Sub-population evolution.}
\gls{matpg} uses two sub-populations: one third \gls{matpg} agents and two thirds independent \gls{maple} agents.
During the evolution of an \gls{matpg} agent, edge destinations can be changed to \gls{maple} agents.
This approach was used in~\cite{maple}, where authors claimed \gls{matpg} alone struggles to learn due to high complexity.
Therefore, the learning is divided with \gls{maple} agents learning behaviors, and \gls{matpg} agents learning how to combine them.

As mentioned in Section~\ref{sec:related_work_gprl}, \gls{matpg} did not surpass \gls{maple} in \cite{maple}.
The hypothesis is that the original \gls{mujoco} benchmark lacks multi-task requirements where \gls{matpg} should excel.
Thus, a \gls{mtrl} application is needed to measure the efficiency of \gls{matpg}.

\section{Use Case: Customized Half Cheetah}
\label{sec:use_case}

To measure \gls{matpg} efficiency, the benchmark must be an \gls{mtrl} environment.
Tasks must be challenging enough to be independent, where mastering one doesn't imply success in others without further training.
They must also be accessible enough for \gls{maple} agents to succeed individually.
In this section, we introduce a new benchmark for \gls{mtrl} in continuous control, using the \textit{Half Cheetah} from the \gls{mujoco} suite. 
Inspired by~\cite{dm_control}, it comprises five distinct obstacle tasks that the \textit{Half Cheetah} must overcome.

The two benchmarks discussed in Section~\ref{sec:related_work_mtrl} fail to meet our requirements.
The \textit{Half Cheetah} body-size variation benchmark exhibits excessive task correlation, as evidenced by the performance of a \gls{maple} agent trained without any variations on \textit{Half Cheetah} and tested on this benchmark, see Figure~\ref{fig:singleBodyPart}.
The benchmark requiring multi-task learning across the \gls{mujoco} suite is unsuitable due to its potential excessive complexity and the need for task-specific, independent knowledge. 
Additionally, unlike real-world multi-task control where tasks involve the same body in different environments, multiple \gls{mujoco} environments introduce distinct input and action spaces, and body morphologies.

Our proposed benchmark strikes a balance: the tasks are independent enough to make achieving a high score challenging for an agent encountering them for the first time, yet they retain similarity through shared \textit{Half Cheetah} behaviors.
This setup also enables evaluation in single episodes featuring sequential obstacles.


\begin{figure*}[!t]
    \centering
    \begin{subfigure}[b]{0.32\textwidth}
        \includegraphics[width=\textwidth]{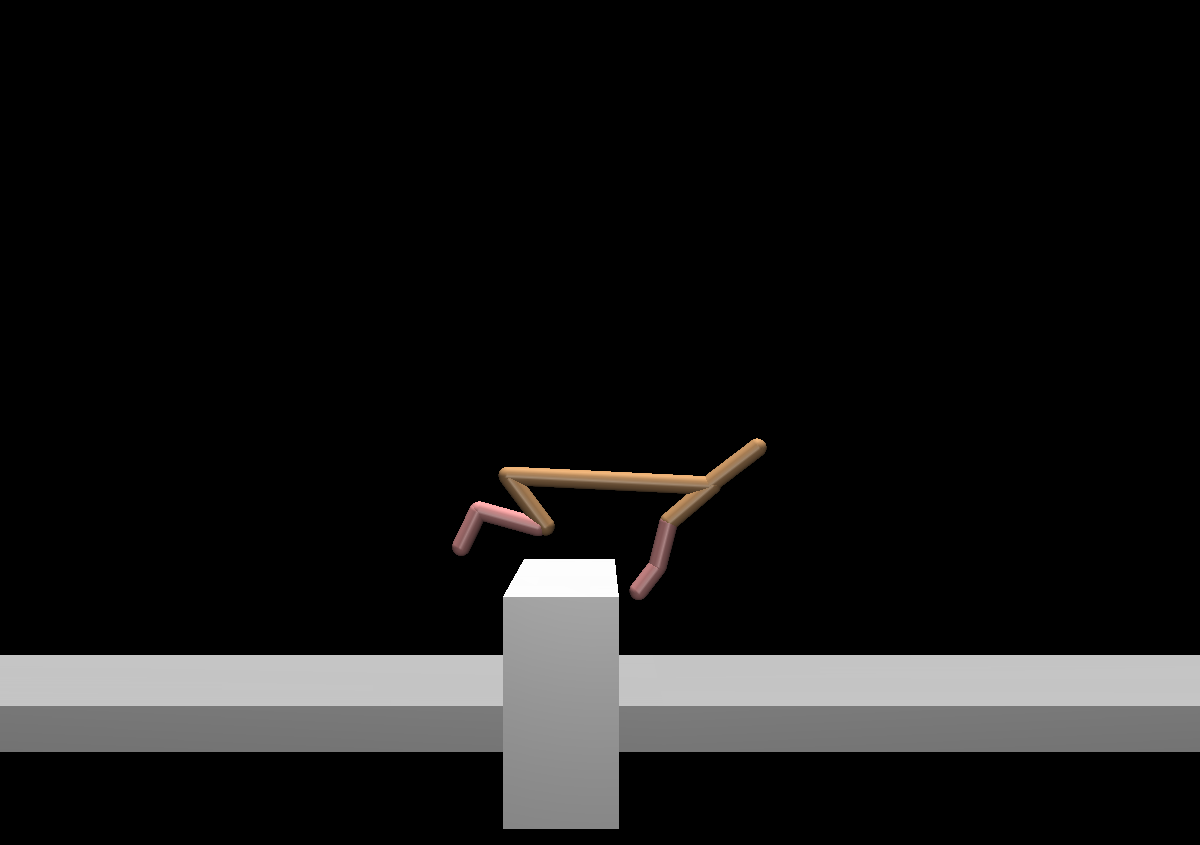}
        \caption{Wall}
    \end{subfigure}
    \begin{subfigure}[b]{0.32\textwidth}
        \includegraphics[width=\textwidth]{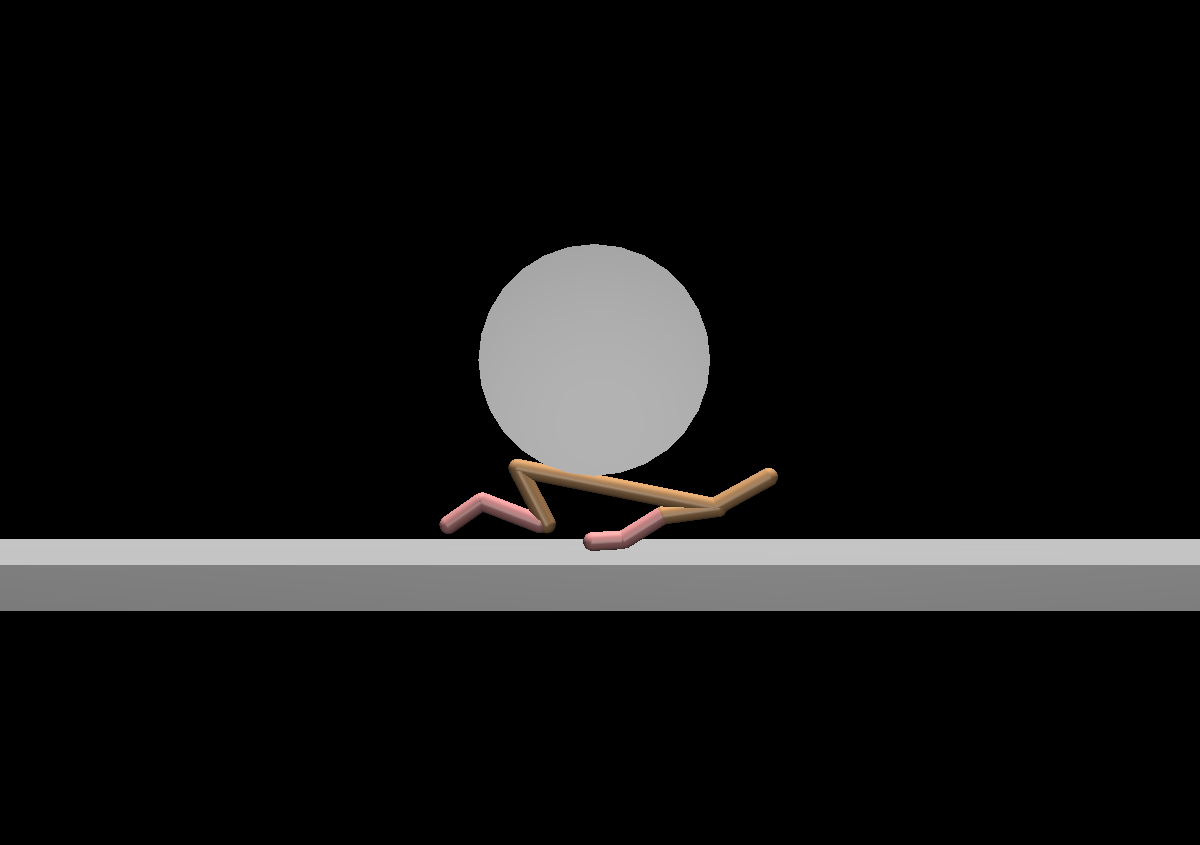}
        \caption{Down}
    \end{subfigure}
    \begin{subfigure}[b]{0.32\textwidth}
        \includegraphics[width=\textwidth]{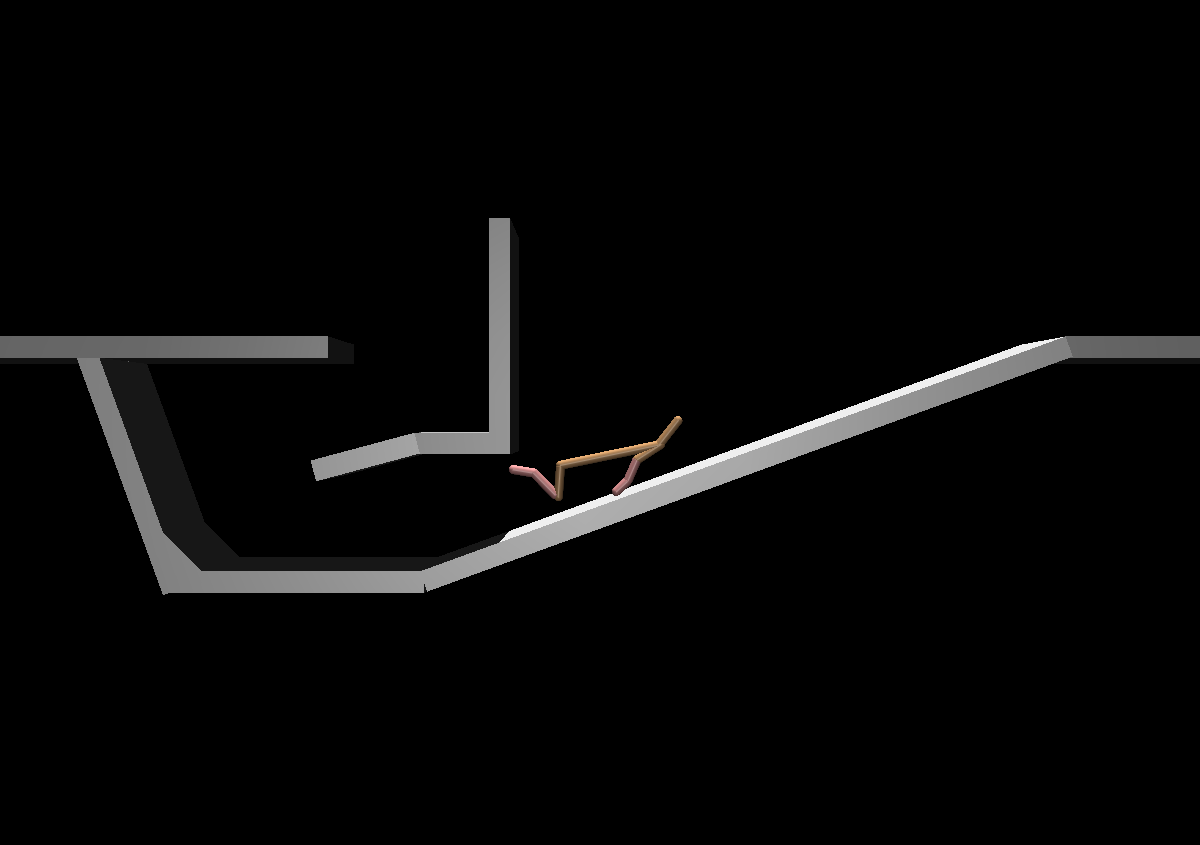}
        \caption{Maze}
    \end{subfigure}
    \begin{subfigure}[b]{0.32\textwidth}
        \includegraphics[width=\textwidth]{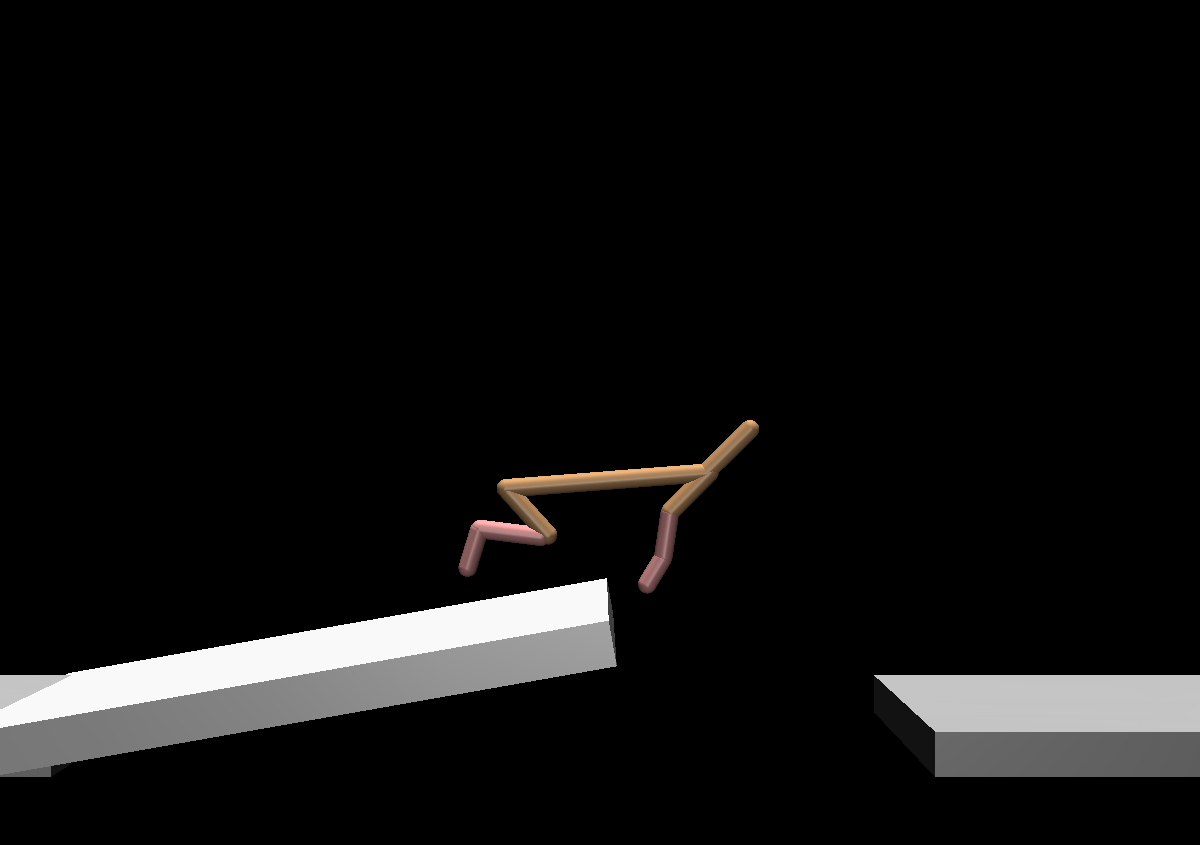}
        \caption{Jump}
    \end{subfigure}
    \begin{subfigure}[b]{0.32\textwidth}
        \includegraphics[width=\textwidth]{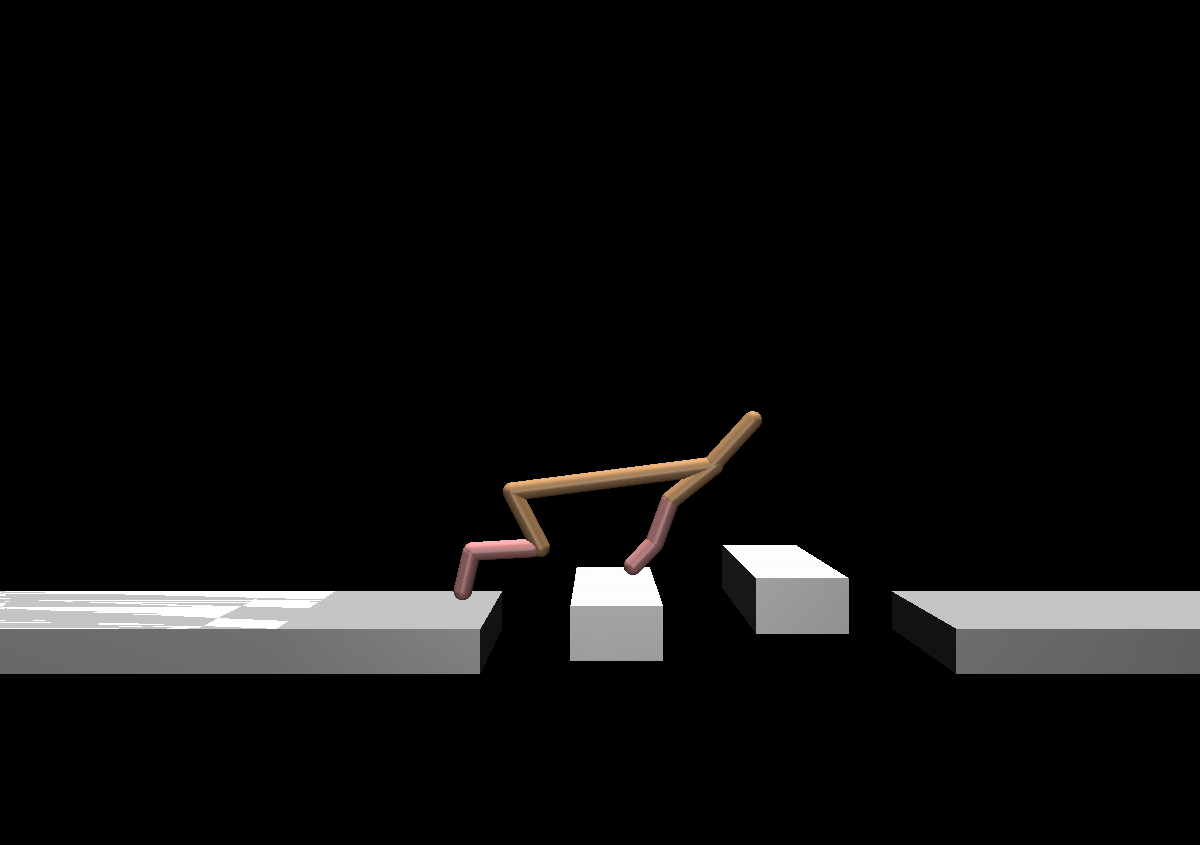}
        \caption{Stairs}
    \end{subfigure}
    \caption{The different obstacles used in the benchmark for the \textit{Half Cheetah}.}
    \label{fig:obstacles}
\end{figure*}

\subsection{The obstacles}
\label{sec:use_case_obstacles}

Figure~\ref{fig:obstacles} shows the different obstacles used in the benchmark:

\begin{itemize}
    \item \textbf{Wall:} A wall that the \textit{Half Cheetah} needs to climb over.
    \item \textbf{Down:} A cylinder that creates a small tunnel through which the \textit{Half Cheetah} can only pass by getting down.
    \item \textbf{Maze:} A maze where the \textit{Half Cheetah} needs to fall, then go backward, and finally climb a slope. The objective is not the slope itself, so the friction on the slope is maximized to help the \textit{Half Cheetah}.
    \item \textbf{Jump:} As with the previous obstacle, the objective is the jump itself, so the friction on the small slope beforehand is increased to help the \textit{Half Cheetah} gain speed.
    \item \textbf{Stairs:} A set of stairs with holes that the \textit{Half Cheetah} needs to avoid.
\end{itemize}

\subsection{The environment}
\label{sec:use_case_environment}

The environment is based on the \textit{Half Cheetah-v5} from Gymnasium~\cite{gymnasium}.
In this environment, the agent performs six actions, applying torques to each of the six joints of the \textit{Half Cheetah}, as shown in Figure~\ref{fig:half_cheetah}.
To learn these actions, the sensor inputs are composed of 17 values representing position and velocity information. 
The default length of an episode is 1000 steps.
The objective is to move the \textit{Half Cheetah} as far to the right as possible.

\begin{figure}[t]
  \begin{center}
    \includegraphics[width=1\textwidth]{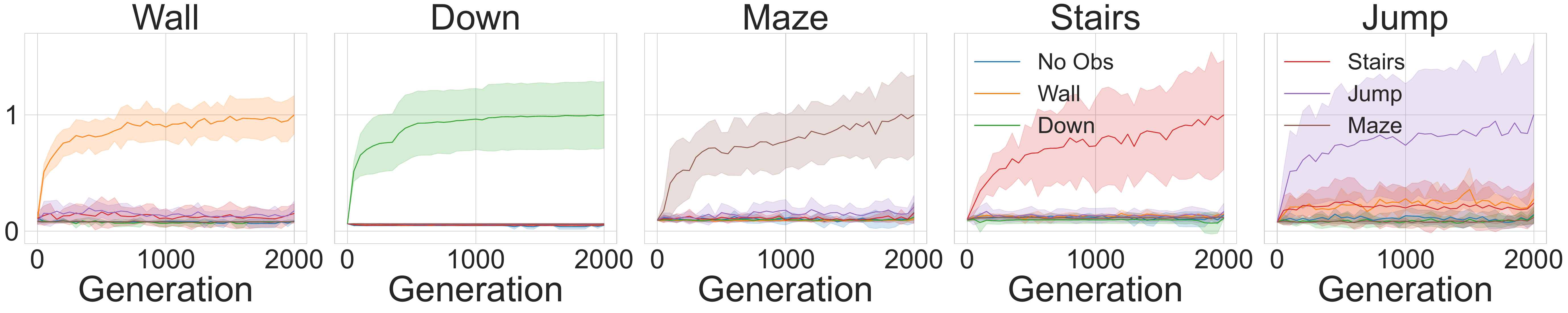}
  \end{center}
  \caption{Validation score achieve on every obstacle, by populations of \gls{maple} agents trained on each obstacle individually and one on no obstacle. The main curve is the mean score over five seeds, and the shaded area is the standard deviation.}
  \label{fig:singleObstacle}
\end{figure}

\begin{wrapfigure}{R}{0.5\textwidth}
\vspace{-10pt}
  \centering
  \includegraphics[width=0.48\textwidth]{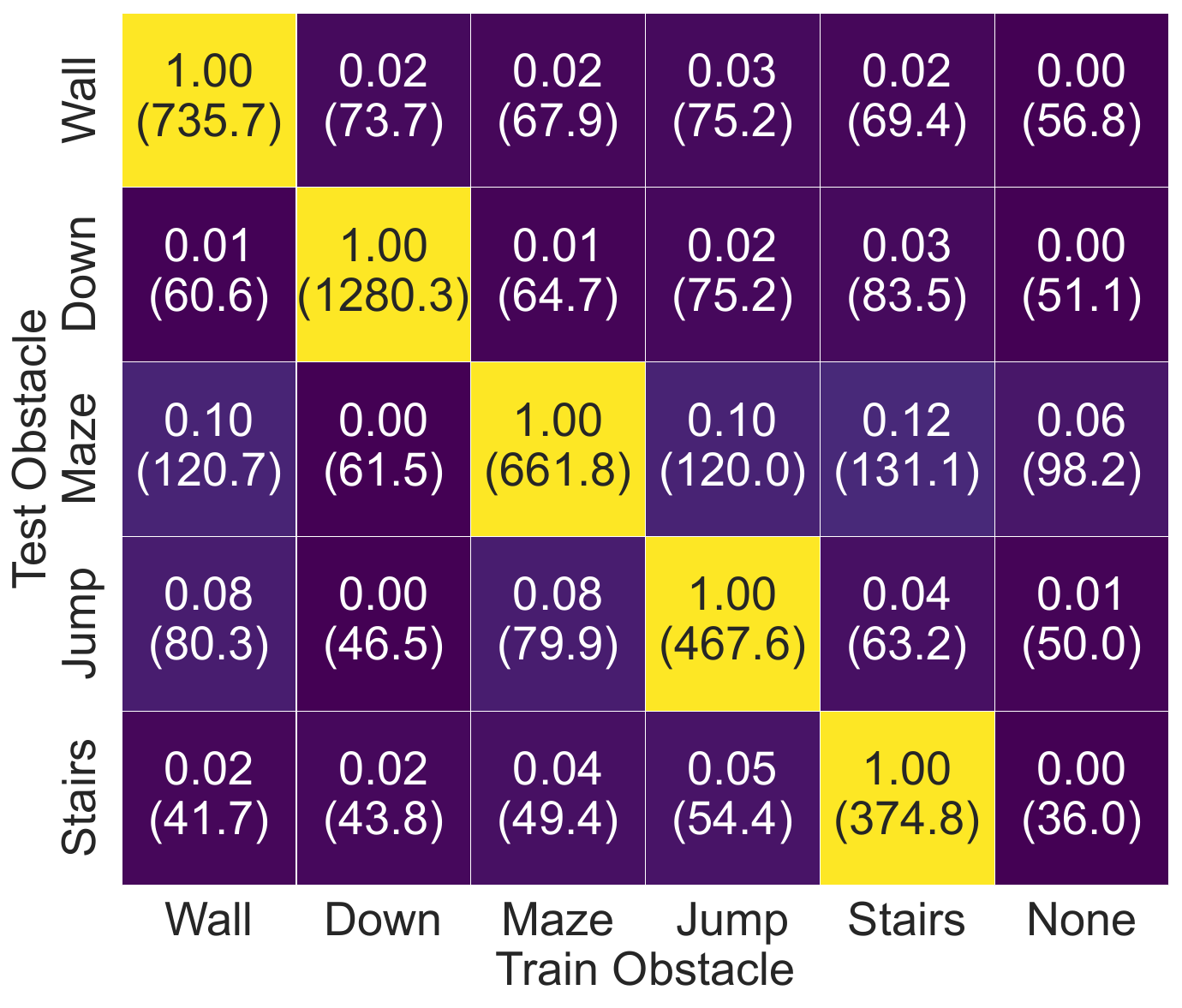} 
  \caption{Correlation matrix of the nomalized score. The x-axis is the obstacle the \gls{maple} agents have been trained on, while the y-axis is the normalized scores reached on each obstacle. The normalized score presented is obtained by dividing the score of the cell by the score of the trained obstacle, the original score is shown within the brackets}
  \label{fig:singleCorrelation}
\vspace{-6pt}
\end{wrapfigure}

To implement obstacles, the environment is divided into 10-meter sections.
When the \textit{Half Cheetah} enters a section, the center of the obstacle is randomly placed between the fourth and fifth meter.
This randomness of the position is important to ensure no overfitting.
Two observations are added to the standard 17: the identifier of the current obstacle and its center position.
While in the obstacle section, the distance to this center is provided, even if negative.

Two additional modifications from the original MuJoCo task were made.
First, to prevent the \textit{Half Cheetah} from falling on its back, the episode ends if the z-orientation of the torso leaves $[-2.5,\ 2.5]$, to avoid undesirable behavior, particularly in the \textit{Maze}.
Second change is on the reward, originally divided in two sub rewards.
The forward reward, rewarding the agent for going as fast as possible to the right.
The control cost, penalizing the agent for applying high torque to the agent motors.
While control cost makes sense for pure speed maximization, it becomes less necessary here and can create strong local minima if tasks aren't learned quickly, thus only the forward reward is used.

\subsection{Initial results}
\label{sec:use_case_maple}

Based on the experimental setup detailed later in Section~\ref{sec:experimental_setup_param}, \gls{maple} agents are trained on each obstacle individually using five seeds. 
Figure~\ref{fig:singleObstacle} presents the results for each task. 
At each validation step, agents are tested on all tasks to assess untrained learning.
The results clearly show they do not learn untrained tasks.
This observation also holds for populations trained without obstacles.

To further validate, the best agents from each training run are tested ten times on each obstacle. 
The correlation matrix in Figure~\ref{fig:singleCorrelation} summarizes these results.
Since obstacles like \textit{Maze} require more time steps than \textit{Jump}, raw scores aren't directly comparable.
The matrix uses normalized scores, dividing each score by the average achieved by populations trained on that specific obstacle.
The correlation matrix reveals complete task independence: agents fail on unseen tasks.
This contrasts with Figure~\ref{fig:singleBodyPart}, where such generalization does occur.

\begin{wrapfigure}{R}{0.5\textwidth}
\vspace{-25pt}
  \centering
  \includegraphics[width=0.48\textwidth]{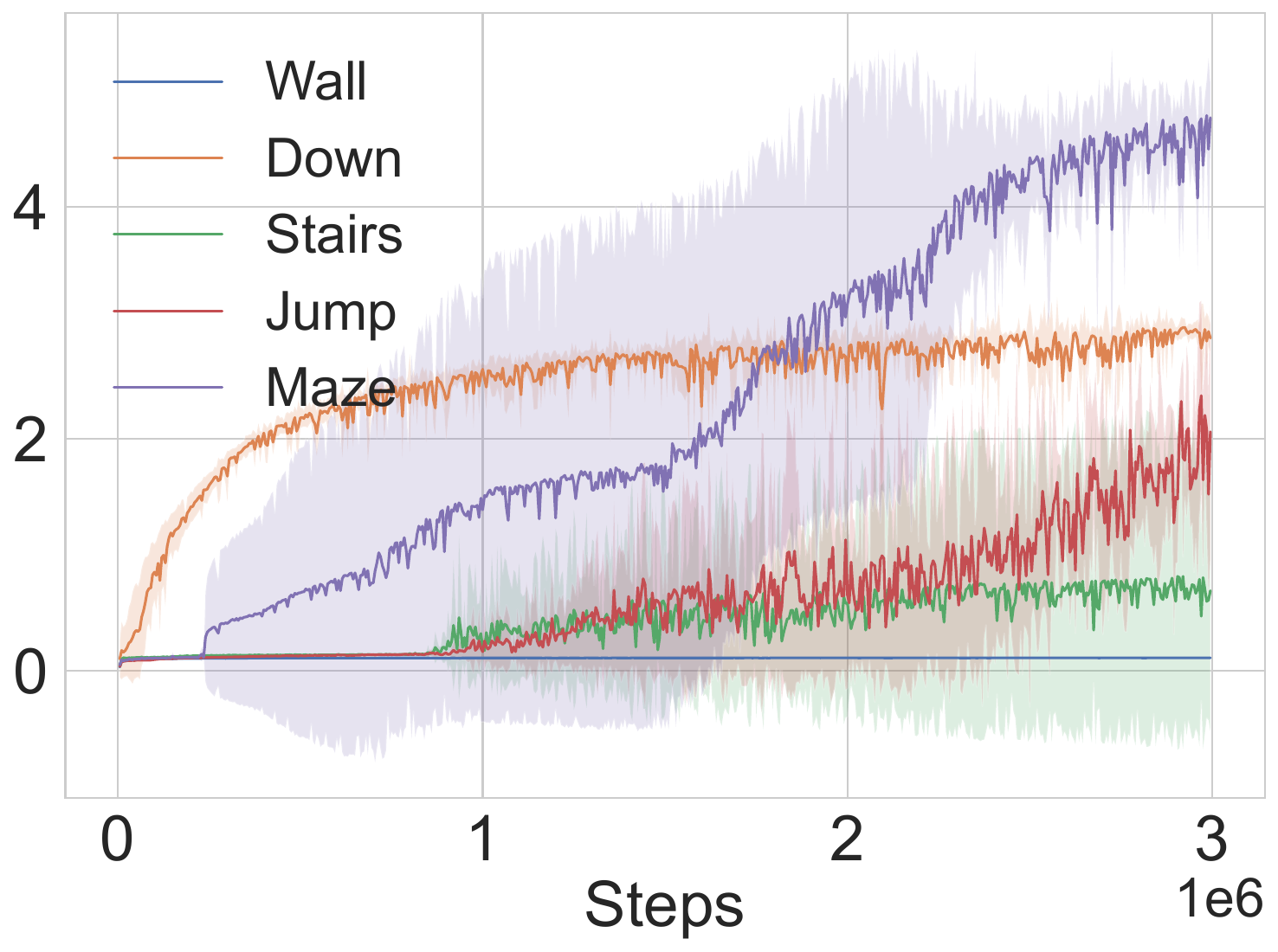} 
  \caption{Validation score achieve on every obstacle, by the \gls{sac}. The main curve is the mean score over five seeds, and the shaded area is the standard deviation.}
  \label{fig:sacResults}
\vspace{-30pt}
\end{wrapfigure}

Thanks to this training, a coefficient can be calculated for each obstacle to indicate its level of difficulty.
Since each task is independent, the difficulty cannot be measure solely by the final score.
It is more relevant to base the difficulty on how quickly the task can be learned.
Additionally, a higher instability between seeds indicates a higher difficulty.
The metric is based on the \gls{auc} of the training~\cite{auc}, corresponding to the following equation:
\begin{equation}
\text{For each task } t, \quad \text{AUC}_t = \frac{\sum_{i=1}^{n}(\mu_{i,t} - \sigma_{i,t})}{\mu_{n,t}}
\end{equation}
Where $n$ is the number of generations, $\mu_{i,t}$ and $\sigma_{i,t}$ represent the average and standard deviation of the score at generation $i$ for task $t$.
The \gls{auc} corresponds to the area under the curve below the shaded region in Figure~\ref{fig:singleObstacle}.
For obstacles \textit{Wall}, \textit{Down}, \textit{Maze}, \textit{Jump}, and \textit{Stairs}, the \gls{auc} values are \( 29.7 \), \( 26.0 \), \( 20.6 \), \( 16.3 \), and \( 10.4 \), respectively.
Normalizing by the highest score yields coefficients of \( 1.00 \), \( 0.88 \), \( 0.69 \), \( 0.55 \), and \( 0.35 \), where higher values indicate easier tasks.
This ranking is supported by one seed failing entirely on the \textit{Stairs} obstacle.

In addition to training with \gls{maple}, experiments used the state-of-the-art \gls{drl} algorithm \gls{sac}~\cite{sac} on the same obstacles.  
Each obstacle was trained with five independent repetitions, with results shown in Figure~\ref{fig:sacResults}, normalized by \gls{maple} scores .  
\gls{sac} consistently learned the \textit{Down} obstacle quickly.  
\textit{Maze} and \textit{Stairs} generally required over one million steps, except one \textit{Maze} seed learning faster.  
For \textit{Jump}, only one seed succeeded, and none on \textit{Wall}.  

Although \gls{sac} learns efficient behaviors for \textit{Down} and \textit{Maze}, it struggles with new strategies needed for obstacles like \textit{Wall}.  
While not designed for \gls{mtrl}, when trained across all obstacles simultaneously, it solved only two versus three when trained separately, supporting the independence of the obstacles.

\section{Experimental Setup}
\label{sec:experimental_setup}

This section describes the experimental setup of the study.
The parametrization mainly follows~\cite{maple}.
Section~\ref{sec:experimental_setup_selection} focuses on selection methods, the main change from the original \gls{matpg} paper, presenting both tournament and lexicase selection.
The parametrization is then detailed in Section~\ref{sec:experimental_setup_param}.

\subsection{Selection methods}
\label{sec:experimental_setup_selection}

\subsubsection{Tournament selection~\cite{tournament}} is the algorithm originally used in \gls{matpg}.
It is widely used and particularly suited for single-task environments.
The process for each generation is as follows:
\begin{itemize}
    \item A small proportion of top-performing agents, called \textit{elites}, are copied directly to the next generation.
    \item The remaining population is randomly divided into fixed-size subsets. From each subset, the best agent, the winner is selected for reproduction.
    \item Winners undergo mutation and crossover to generate offspring.
    \item The next generation consists of elites and the new offspring.
\end{itemize}

\subsubsection{Lexicase selection~\cite{lexicase}} was originally designed for classification tasks to promote agents capable of learning multiple classes.
The process to select each survivor is:
\begin{itemize}
    \item A random task order is generated.
    \item For each task in order, only the top-performing agents are retained.
    \item After all tasks, if one agent remains, it is selected. If multiple remain, one is chosen randomly.
\end{itemize}
This repeats until the desired number of survivors is achieved.
Survivors generate offspring for the next generation, while others are discarded.

Lexicase selection was proposed for discrete integer scores where top performers often tie.
Epsilon-lexicase~\cite{lexicaseEpsilon} is a variant for regression problems with continuous scores, where exact equality is rare.
In epsilon-lexicase, selection is relaxed: all agents with scores in $[ \text{max}(score) \times \epsilon,\ \text{max}(score) ]$ are retained, where $\epsilon$ is computed from median absolute deviation.
This allows more flexible selection.
In our context, epsilon-lexicase is more appropriate due to continuous scores.
However, since scores are very high in \gls{rl}, we multiply the epsilon coefficient by $0.1$ to limit its impact.

\subsection{Parametrization}
\label{sec:experimental_setup_param}

To achieve the same solution as in~\cite{maple}, we use the same training framework based on the open source Gegelati library~\cite{gegelati}.
This C++ library originally implemented \gls{tpg} and was extended for \gls{maple}.

We use the same hyperparameters as~\cite{maple} for \gls{matpg} training.
The instruction set is $H = \{ +$, $-$, $\times$, $\div$, $\max$, $\exp$, $\log$, $\sin$, $\cos$, $\operatorname{tan}$, $\operatorname{modulo} \}$.
Instructions include constants multiplied by their results, with constants mutation based on the mutation in AutoML-Zero~\cite{automlzero}.
For \gls{maple} agents on single tasks in Section~\ref{sec:use_case_maple}, identical hyperparameters with tournament selection are used: 1000 agents and \gls{maple} proportion of 1.
For \gls{matpg}, values are set to 1500 agents with 0.67 \gls{maple} proportion, leading to 1000 \gls{maple} and 500 \gls{matpg} agents.

The number of episodes per generation indicates how many times the process needs to be repeated to achieve an accurate score.
A value of three per generation is chosen for training, leading to fifteen episodes when five obstacles are used.

\section{Experiments}
\label{sec:experiments}

In this Section, the evaluation of \gls{matpg} is done based on the experimental setup established in Section~\ref{sec:experimental_setup}.
An ablation study is done to evaluate the performances. 
In a way, \gls{maple} is a simplification of \gls{matpg}, thus both algorithms are tested with either tournament and lexicase selection.
Videos showing the behaviors are available in the supplementary materials.

We show through an ablation study the results of MATPG compared to MAPLE, with either the lexicase or the tournament selection.
In addition to evaluating the scalability of the different solutions in overcoming obstacles, the tests are conducted not only on the full set of five obstacles but also on subsets comprising two, three, and four obstacles. 
The selection of tasks is based on their difficulty, as computed in Section~\ref{sec:use_case_maple}. Specifically, for the two-obstacle scenario, the two easiest obstacles are used for training. 
This approach is incrementally extended up to the full set of five obstacles.

\begin{figure*}[!t]
    \centering
    \begin{subfigure}[b]{0.99\textwidth}
        \includegraphics[width=\textwidth]{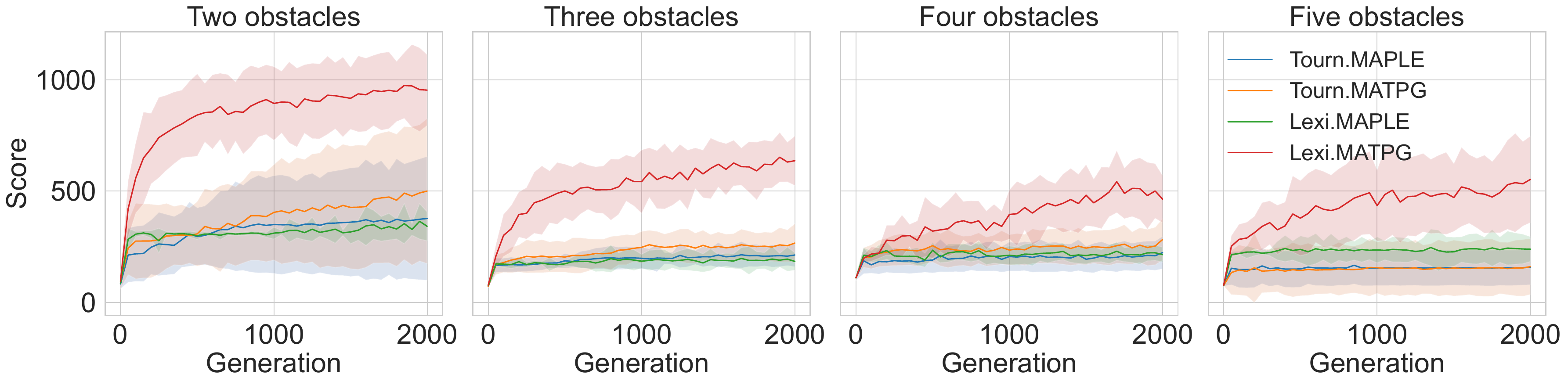}
        \caption{Results over generations. The main curve is the mean score over ten seeds, and the shaded area is the standard deviation.}
        \label{fig:multiObstacleResults_training}
    \end{subfigure}
    \begin{subfigure}[b]{0.99\textwidth}
        \includegraphics[width=\textwidth]{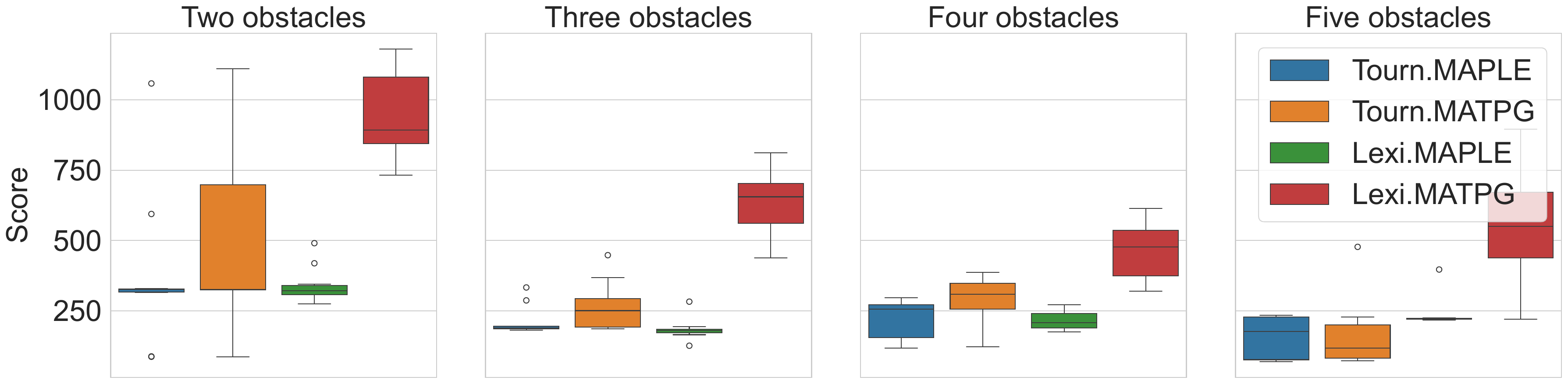}
        \caption{Box plots showing the performances of the best agent of each seed for each solution at the end of the training.}
        \label{fig:multiObstacleResults_validation}
    \end{subfigure}
    \caption{Validation score achieve on average on five episodes with all trained obstacles showing in random order. Tourn. and Lexi. respectively mean tournament and lexicase selection. }
    \label{fig:multiObstacleResults}
\end{figure*}

Figure~\ref{fig:multiObstacleResults} presents the training performance of the four evaluated approaches across environments with two to five obstacles. 
In all scenarios, \gls{matpg} combined with lexicase selection consistently outperforms both \gls{matpg} with tournament selection and \gls{maple} under either selection method. 
For the two-obstacle environment, \gls{matpg} with tournament selection shows slightly better performance than \gls{maple}, though the difference is not statistically significant. 
As the number of obstacles increases, the performance of \gls{matpg} with tournament selection aligns closely with that of \gls{maple}. 
These results indicate that the original \gls{matpg} approach~\cite{maple} alone struggles to solve \gls{mtrl} tasks with independent objectives. 
However, integrating lexicase selection enables \gls{matpg} to effectively perform multi-task learning in continuous control environments.

\begin{table}[ht]
\caption{Best normalized score on each obstacle, when trained for five obstacles.
         The normalized score is obtained by dividing the score by the average score obtained by the single obstacle training in Section~\ref{sec:use_case_maple}. The score presented is the average score among ten seeds plus or minus the standard deviation.}
\centering
\label{tab:multiObstaclePerTasks}
\begin{tabular}{l|l||l|l|l|l|l}
\toprule
\textbf{Algo.} & \textbf{Selection} & \textbf{Wall} & \textbf{Down} & \textbf{Maze} & \textbf{Jump} & \textbf{Stairs} \\
\midrule
\multicolumn{7}{c}{\textbf{Best score by any agent within the whole population}} \\
\midrule
MAPLE & Tourn. & 0.39±(0.3) & 0.27±(0.2) & 0.18±(0.2) & 0.21±(0.1) & 0.36±(0.3) \\
MATPG & Tourn. & 0.60±(0.3) & 0.17±(0.2) & 0.22±(0.2) & 0.23±(0.1) & 0.63±(0.3) \\
MAPLE & Lexi. & 0.73±(0.2) & 0.72±(0.2) & \textbf{0.70±(0.2)} & 0.22±(0.0) & 0.79±(0.2) \\
MATPG & Lexi. & \textbf{0.81±(0.1)} & \textbf{0.94±(0.2)} & 0.69±(0.2) & \textbf{0.31±(0.1)} & \textbf{0.99±(0.3)} \\
\midrule
\multicolumn{7}{c}{\textbf{Score by the best agent on episodes with all obstacles}} \\
\midrule
MAPLE & Tourn. & 0.24±(0.2) & 0.25±(0.2) & 0.12±(0.1) & 0.13±(0.1) & 0.21±(0.2) \\
MATPG & Tourn. & 0.30±(0.2) & 0.15±(0.2) & 0.13±(0.2) & 0.12±(0.1) & 0.30±(0.3) \\
MAPLE & Lexi. & 0.12±(0.2) & 0.60±(0.2) & 0.15±(0.2) & 0.10±(0.0) & 0.10±(0.1) \\
MATPG & Lexi. & \textbf{0.46±(0.3)} & \textbf{0.91±(0.2)} & \textbf{0.24±(0.3)} & \textbf{0.14±(0.1)} & \textbf{0.45±(0.4)} \\
\bottomrule
\end{tabular}
\end{table}

\begin{wraptable}{R}{0.48\textwidth}
\vspace{-10pt}
\caption{Welch's $t$-test results. $\alpha_{\text{corr}} = 0.005$ with Bonferroni correction ($N=10$). Cohen's $d > 2$ indicates a very large effect.}
\label{tab:significance}
\centering
\begin{tabular}{l|c|c}
 & $p$-value & Cohen's $d$ \\
\midrule
Tourn. MAPLE  & 0.000 & 2.68 \\
Tourn. MATPG  & 0.000 & 2.40 \\
Lexi. MAPLE   & 0.001 & 2.20 \\
\end{tabular}
\vspace{-6pt}
\end{wraptable}

Statistical significance is assessed following prior work \cite{significance}, using a two-tailed Welch’s $t$-test \cite{welch}, which is appropriate for small samples with unequal variances.
With $N=10$ independent runs per method, we apply a Bonferroni correction to control the family-wise error rate, yielding an adjusted significance threshold of $\alpha_{\text{corr}} = 0.005$.
Practical significance is measured using Cohen’s $d$, and, we consider $d > 2$ to indicate a sufficiently large effect under limited-run settings.
The results in Table~\ref{tab:significance} report the $p$-values and corresponding Cohen’s $d$ for Lexicase \gls{matpg} against all solutions.
Across all comparisons, we observe $p < 0.005$ and $d > 2$, demonstrating statistically significant and practically substantial improvements.
Therefore, the performance gains are unlikely to be due to random variation.

Table~\ref{tab:multiObstaclePerTasks} presents the results obtained by each solution for individual obstacles. 
The first part of the table reports, for each training, the best score achieved by any agent within its entire population, that is, each score can be given by a distinct agent within the population that specialized in the task. 
Here, both \gls{matpg} and \gls{maple} perform significantly better when using lexicase selection compared to tournament selection, highlighting the ability of lexicase selection to naturally produce elite agents for each task. 
Notably, \gls{matpg} consistently achieves equal or higher scores per obstacle than \gls{maple} when lexicase selection is employed.

The second part of the table evaluates the best agent from each population on the individual obstacles. 
The agent is selected based on performance across episodes containing all obstacles. 
In an ideal scenario, \gls{matpg} with lexicase selection would achieve the same results in both parts of the table. 
The observed discrepancy indicates that the method does not fully exploit the sub-population generated by \gls{maple}. 
Nevertheless, even under this non-optimal condition, \gls{matpg} with lexicase selection outperforms all other approaches, including \gls{maple} with lexicase selection, across every obstacle. 
These findings suggest that lexicase selection, regardless of the underlying algorithm, can produce elite agents capable of mastering individual tasks, whereas \gls{matpg} is necessary to consolidate this ability into a single agent capable of solving multiple obstacles simultaneously.

\section{Interpretability of the best agent}
\label{sec:interpretability}

In this section, the interpretability of the best \gls{matpg} agent is analyzed.
This agent achieves high accuracy on the \textit{Wall}, \textit{Down}, and \textit{Stairs} obstacles, with normalized scores of $0.74$, $0.95$, and $0.77$, respectively.  
However, its performance decreases on the \textit{Maze} and \textit{Jump} obstacles, where it reaches scores of $0.32$ and $0.16$.  

\begin{figure}[t]
  \begin{center}
    \includegraphics[width=0.8\textwidth]{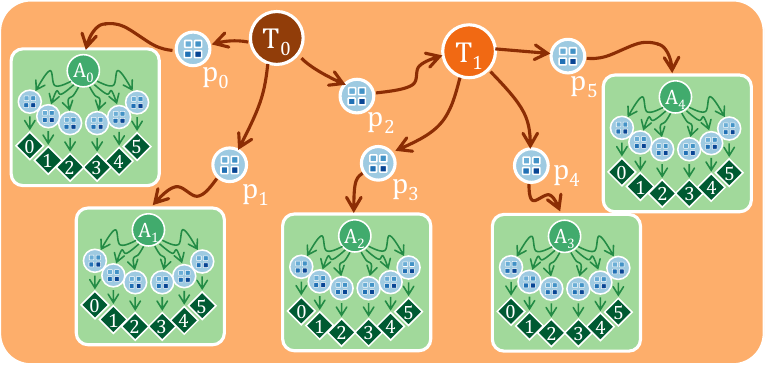}
  \end{center}
  \caption{Topology of the best \gls{matpg} agent on the five obstacles.}
  \label{fig:matpg_best}
\end{figure}

To better understand its behavior, the topology of the agent is illustrated in Figure~\ref{fig:matpg_best}.
The agent consists of two teams vertices, each containing three programs, and five action vertices, where each action vertex includes six programs.  
Five of the programs connect to the different action vertices, while the remaining connect the two teams.  
The team vertices, action vertices, and programs are labeled in Figure~\ref{fig:matpg_best}, and the detailed behavior of the context programs is described in Algorithm~\ref{algo:matpg_best_program}. The activation behavior of the agent in response to specific obstacle types can be summarized as follows:

\begin{itemize}
    \item \textbf{Jump} and \textbf{Stairs}: Under these conditions, program \textit{$p_0$} consistently produces the highest output value, resulting in the activation of action vertex \textit{$A_0$}.
    \item \textbf{Wall}: When encountering a wall obstacle, program \textit{$p_1$} dominates, thereby activating action vertex \textit{$A_1$}.
    \item \textbf{Down} and \textbf{Maze}: For these obstacle types, program \textit{$p_2$} yields the highest output, leading to the activation of team vertex \textit{$T_1$}.
    \item \textbf{Maze (within $T_1$)}: Once the team vertex \textit{$T_1$} is activated, encountering a maze obstacle causes program \textit{$p_4$} to produce the maximum output, which in turn activates action vertex \textit{$A_3$}.
    \item \textbf{Down (within $T_1$)}: This obstacle exhibits a more complex dynamic. When the agent is positioned far from the obstacle center, programs \textit{$p_3$} and \textit{$p_4$} alternately dominate, producing a cyclic activation pattern lasting between two and four steps per program. As the agent approaches the center of the obstacle, program \textit{$p_5$} becomes predominant, activating action vertex \textit{$A_4$}.
\end{itemize}

In summary, the \textit{Jump} and \textit{Stairs} obstacles are handled by \textit{$A_0$}, the \textit{Wall} by \textit{$A_1$}, and the \textit{Maze} by \textit{$A_3$}.  
The \textit{Down} obstacle is addressed collaboratively by \textit{$A_2$}, \textit{$A_3$}, and \textit{$A_4$}.
The instructions used by each program fully explain this.
In short, the traversal paths are mostly due to the state variables $s_{17}$ and $s_{18}$, which correspond respectively to the obstacle identifier index and the distance to the obstacle, making the following analysis deterministic.

\begin{algorithm}[H]
\caption{Programs used by the teams of the best \gls{matpg} agent, see Figure~\ref{fig:matpg_best}. Each line shows the Program $p_i$ that is computed, depending on the state inputs $s_j$. The agent only used six inputs from the nineteen of the state inputs. $s_3$, $s_6$, $s_7$ and $s_8$ are inputs respectively corresponding to the angle of the back shin, the angle of the front shin, the angle of the front foot and the velocity of the x-coordinate of the front tip. $s_{17}$ and $s_{18}$ are additional inputs corresponding respectively to the identifier index of the obstacle and the distance to the center of the obstacle.}
\label{algo:matpg_best_program}
\begin{algorithmic}[0]
\State $p_0 \gets \tan(-0.751904 \times (s_{17} + s_6))$
\State $p_1 \gets -7.19471 \times \cos(-0.114644 \times s_8) - 6.002294 \times \frac{s_7}{s_{17}}$
\State $p_2 \gets -0.064495 \times \log(-0.416072 \times \sin(-0.855615 \times s_6))$
\State $p_3 \gets \tan((s_3 - \tan(s_{17}))  \times 0.727467 )$
\State $p_4 \gets \tan(0.727467 \times (\tan(0.583892 \times \cos(-0.719604 \times s_{18})) + \tan(s_{17})))$
\State $p_5 \gets \sin(0.064883 \times \cos(s_{18}))$
\end{algorithmic}
\end{algorithm}

\section{Discussion and conclusion}
\label{sec:conclusion}

In this work, we designed a new \gls{mtrl} benchmark to assess \gls{matpg} efficiency compared to \gls{maple}.
Built on the \textit{Half Cheetah} environment from \gls{mujoco}, it features five distinct obstacles that the agent must overcome.
These tasks are designed to satisfy two key properties.
First, each task is mutually independent, ensuring that solving one or the original locomotion task does not provide sufficient information to solve the others.
Second, all tasks remain learnable by a single \gls{maple} agent, establishing a baseline that verifies feasibility without requiring specialized methods.
Finally, the benchmark is readily extensible, allowing future work to incorporate additional obstacles depending on experimental needs.
A current limitation is that agents lack explicit obstacle structure information.
Future work could investigate richer task representations by incorporating structural features to enable knowledge sharing.

We evaluate both  \gls{maple} and \gls{matpg} under tournament selection, their default configuration, and lexicase selection, which promotes task specialization.
Lexicase selection produces populations with improved coverage of task-specific behaviors in both algorithms.
However, while \gls{maple} with lexicase evolves agents solving tasks independently, only \gls{matpg} with lexicase achieves significantly higher performance when a single agent faces all obstacles simultaneously.
Analysis shows \gls{matpg} preserves the inherent interpretability of \gls{gp} with clear and human-readable decision paths and obstacle-specific behaviors.
Although lexicase selection performs better, it requires prior task knowledge.
Developing evolutionary techniques without such requirements represents a promising future direction.

\section*{Acknowledgements}
This research was funded, in whole or in part, by the Agence Nationale de la Recherche (ANR), grant ANR-22-CE25-0005-01. A CC BY license is applied to the AAM resulting from this submission, in accordance with  the open access conditions of the grant.

\bibliographystyle{splncs04}
\bibliography{reference}

@misc{mujocoMT,
      title={Knowledge Transfer in Multi-Task Deep Reinforcement Learning for Continuous Control}, 
      author={Zhiyuan Xu and Kun Wu and Zhengping Che and Jian Tang and Jieping Ye},
      year={2020},
      eprint={2010.07494},
      archivePrefix={arXiv},
      primaryClass={cs.LG},
      url={https://arxiv.org/abs/2010.07494}, 
}

@article{dqn,
  title={Human-level control through deep reinforcement learning},
  author={Mnih, Volodymyr and Kavukcuoglu, Koray and Silver, David and Rusu, Andrei A and Veness, Joel and Bellemare, Marc G and Graves, Alex and Riedmiller, Martin and Fidjeland, Andreas K and Ostrovski, Georg and others},
  journal={nature},
  volume={518},
  number={7540},
  pages={529--533},
  year={2015},
  publisher={Nature Publishing Group}
}

@article{gymnasium,
  title={Gymnasium: A Standard Interface for Reinforcement Learning Environments},
  author={Towers, Mark and Kwiatkowski, Ariel and Terry, Jordan and Balis, John U and De Cola, Gianluca and Deleu, Tristan and Goul{\~a}o, Manuel and Kallinteris, Andreas and Krimmel, Markus and KG, Arjun and others},
  journal={arXiv preprint arXiv:2407.17032},
  year={2024}
}

@INPROCEEDINGS{mujoco,
  author={Todorov, Emanuel and Erez, Tom and Tassa, Yuval},
  booktitle={2012 IEEE/RSJ International Conference on Intelligent Robots and Systems}, 
  title={MuJoCo: A physics engine for model-based control}, 
  year={2012},
  volume={},
  number={},
  pages={5026-5033},
  doi={10.1109/IROS.2012.6386109}
}

@misc{sac,
      title={Soft Actor-Critic: Off-Policy Maximum Entropy Deep Reinforcement Learning with a Stochastic Actor}, 
      author={Tuomas Haarnoja and Aurick Zhou and Pieter Abbeel and Sergey Levine},
      year={2018},
      eprint={1801.01290},
      archivePrefix={arXiv},
      primaryClass={cs.LG},
      url={https://arxiv.org/abs/1801.01290}, 
}

@misc{ppo,
      title={Proximal Policy Optimization Algorithms}, 
      author={John Schulman and Filip Wolski and Prafulla Dhariwal and Alec Radford and Oleg Klimov},
      year={2017},
      eprint={1707.06347},
      archivePrefix={arXiv},
      primaryClass={cs.LG},
      url={https://arxiv.org/abs/1707.06347}, 
}

@misc{td3,
      title={Addressing Function Approximation Error in Actor-Critic Methods}, 
      author={Scott Fujimoto and Herke van Hoof and David Meger},
      year={2018},
      eprint={1802.09477},
      archivePrefix={arXiv},
      primaryClass={cs.AI},
      url={https://arxiv.org/abs/1802.09477}, 
}

@article{ale,
   title={The Arcade Learning Environment: An Evaluation Platform for General Agents},
   volume={47},
   ISSN={1076-9757},
   url={http://dx.doi.org/10.1613/jair.3912},
   DOI={10.1613/jair.3912},
   journal={Journal of Artificial Intelligence Research},
   publisher={AI Access Foundation},
   author={Bellemare, M. G. and Naddaf, Y. and Veness, J. and Bowling, M.},
   year={2013},
   month=jun, pages={253–279}
}

@misc{impala,
      title={IMPALA: Scalable Distributed Deep-RL with Importance Weighted Actor-Learner Architectures}, 
      author={Lasse Espeholt and Hubert Soyer and Remi Munos and Karen Simonyan and Volodymir Mnih and Tom Ward and Yotam Doron and Vlad Firoiu and Tim Harley and Iain Dunning and Shane Legg and Koray Kavukcuoglu},
      year={2018},
      eprint={1802.01561},
      archivePrefix={arXiv},
      primaryClass={cs.LG},
      url={https://arxiv.org/abs/1802.01561}, 
}

@article{explainableRL,
  title={Explainable reinforcement learning: A survey and comparative review},
  author={Milani, Stephanie and Topin, Nicholay and Veloso, Manuela and Fang, Fei},
  journal={ACM Computing Surveys},
  volume={56},
  number={7},
  pages={1--36},
  year={2024},
  publisher={ACM New York, NY}
}

@inproceedings{vacherECTA,
  title={Hybrid Genetic Programming and Deep Reinforcement Learning for Low-complexity Robot Arm Trajectory Planning},
  author={Vacher, Quentin and Beuve, Nicolas and Allaire, Paul and Marty, Thibaut and Dardaillon, Micka{\"e}l and Desnos, Karol},
  booktitle={16th International Conference on Evolutionary Computation Theory and Applications},
  pages={139--150},
  year={2024}
}

@inproceedings{gpKoza,
  title={Automatic programming of robots using genetic programming},
  author={Koza, John R and Rice, James P},
  booktitle={AAAI},
  volume={92},
  pages={194--207},
  year={1992}
}

@inproceedings{de2025evolution,
  title={Evolution of Inherently Interpretable Visual Control Policies},
  author={De La Torre, Camilo and Nadizar, Giorgia and Lavinas, Yuri and Luga, Herv{\'e} and Wilson, Dennis and Cussat-Blanc, Sylvain},
  booktitle={Proceedings of the Genetic and Evolutionary Computation Conference},
  pages={358--367},
  year={2025}
}

@article{dm_control,
  title={dm\_control: Software and tasks for continuous control},
  author={Tunyasuvunakool, Saran and Muldal, Alistair and Doron, Yotam and Liu, Siqi and Bohez, Steven and Merel, Josh and Erez, Tom and Lillicrap, Timothy and Heess, Nicolas and Tassa, Yuval},
  journal={Software Impacts},
  volume={6},
  pages={100022},
  year={2020},
  publisher={Elsevier}
}

@article{mujocoMTBench,
  author       = {Peter Henderson and
                  Wei{-}Di Chang and
                  Florian Shkurti and
                  Johanna Hansen and
                  David Meger and
                  Gregory Dudek},
  title        = {Benchmark Environments for Multitask Learning in Continuous Domains},
  journal      = {CoRR},
  volume       = {abs/1708.04352},
  year         = {2017},
  url          = {http://arxiv.org/abs/1708.04352},
  eprinttype    = {arXiv},
  eprint       = {1708.04352},
  bibsource    = {dblp computer science bibliography, https://dblp.org}
}

@inproceedings{lgpcartpole,
  TITLE = {{Multi-Objective Genetic Programming for Explainable Reinforcement Learning}},
  AUTHOR = {Videau, Mathurin and Ferreira Leite, Alessandro and Teytaud, Olivier and Schoenauer, Marc},
  URL = {https://inria.hal.science/hal-03886307},
  BOOKTITLE = {{Lecture Notes in Computer Science}},
  ADDRESS = {Madrid, Spain},
  PUBLISHER = {{Springer International Publishing}},
  SERIES = {Lecture Notes in Computer Science},
  VOLUME = {13223},
  PAGES = {278-293},
  YEAR = {2022},
  MONTH = Apr,
  DOI = {10.1007/978-3-031-02056-8\_18},
  HAL_ID = {hal-03886307},
  HAL_VERSION = {v1},
}

@book{lgp,
  title={Basic concepts of linear genetic programming},
  author={Brameier, Markus F and Banzhaf, Wolfgang},
  year={2007},
  publisher={Springer}
}

@inproceedings{nadizarMujoco,
  title={Naturally Interpretable Control Policies via Graph-Based Genetic Programming},
  author={Nadizar, Giorgia and Medvet, Eric and Wilson, Dennis G},
  booktitle={European Conference on Genetic Programming (Part of EvoStar)},
  pages={73--89},
  year={2024},
  organization={Springer}
}

@inproceedings{cgp,
  title={Cartesian genetic programming},
  author={Miller, Julian and Turner, Andrew},
  booktitle={Proceedings of the Companion Publication of the 2015 Annual Conference on Genetic and Evolutionary Computation},
  pages={179--198},
  year={2015}
}

@inproceedings{maple,
  title={MAPLE: Multi-Action Programs through Linear Evolution for Continuous Multi-Action Reinforcement Learning},
  author={Vacher, Quentin and Kelly, Stephen and Naqvi, Ali and Beuve, Nicolas and Djavaherpour, Tanya and Dardaillon, Micka{\"e}l and Desnos, Karol},
  booktitle={Proceedings of the Genetic and Evolutionary Computation Conference},
  pages={1062--1071},
  year={2025}
}

@inproceedings{tpg,
  title={Emergent tangled graph representations for Atari game playing agents},
  author={Kelly, Stephen and Heywood, Malcolm I},
  booktitle={Genetic Programming: 20th European Conference, EuroGP 2017, Amsterdam, The Netherlands, April 19-21, 2017, Proceedings 20},
  pages={64--79},
  year={2017},
  organization={Springer}
}

@inproceedings{tpgmt,
  title={Multi-task learning in atari video games with emergent tangled program graphs},
  author={Kelly, Stephen and Heywood, Malcolm I},
  booktitle={Proceedings of the Genetic and Evolutionary Computation Conference},
  pages={195--202},
  year={2017}
}

@inproceedings{heywoodwalker,
  title={Benchmarking genetic programming in a multi-action reinforcement learning locomotion task},
  author={Amaral, Ryan and Ianta, Alexandru and Bayer, Caleidgh and Smith, Robert J and Heywood, Malcolm I},
  booktitle={Proceedings of the Genetic and Evolutionary Computation Conference Companion},
  pages={522--525},
  year={2022}
}

@article{auc,
  title={K-percent Evaluation for Lifelong RL},
  author={Mesbahi, Golnaz and Panahi, Parham Mohammad and Mastikhina, Olya and White, Martha and White, Adam},
  journal={arXiv preprint arXiv:2404.02113},
  year={2024}
}

@inproceedings{tournament,
  title={A review of tournament selection in genetic programming},
  author={Fang, Yongsheng and Li, Jun},
  booktitle={International symposium on intelligence computation and applications},
  pages={181--192},
  year={2010},
  organization={Springer}
}

@article{lexicase,
  title={Solving uncompromising problems with lexicase selection},
  author={Helmuth, Thomas and Spector, Lee and Matheson, James},
  journal={IEEE Transactions on Evolutionary Computation},
  volume={19},
  number={5},
  pages={630--643},
  year={2014},
  publisher={IEEE}
}

@inproceedings{lexicaseEpsilon,
  title={Epsilon-lexicase selection for regression},
  author={La Cava, William and Spector, Lee and Danai, Kourosh},
  booktitle={Proceedings of the Genetic and Evolutionary Computation Conference 2016},
  pages={741--748},
  year={2016}
}

@misc{automlzero,
      title={AutoML-Zero: Evolving Machine Learning Algorithms From Scratch}, 
      author={Esteban Real and Chen Liang and David R. So and Quoc V. Le},
      year={2020},
      eprint={2003.03384},
      archivePrefix={arXiv},
      primaryClass={cs.LG},
      url={https://arxiv.org/abs/2003.03384}, 
}

@inproceedings{gegelati,
  title={Gegelati: Lightweight artificial intelligence through generic and evolvable tangled program graphs},
  author={Desnos, Karol and Sourbier, Nicolas and Raumer, Pierre-Yves and Gesny, Olivier and Pelcat, Maxime},
  booktitle={Workshop on Design and Architectures for Signal and Image Processing (14th edition)},
  pages={35--43},
  year={2021}
}

@article{rlsurvey,
  title={Deep reinforcement learning: A survey},
  author={Wang, Xu and Wang, Sen and Liang, Xingxing and Zhao, Dawei and Huang, Jincai and Xu, Xin and Dai, Bin and Miao, Qiguang},
  journal={IEEE Transactions on Neural Networks and Learning Systems},
  volume={35},
  number={4},
  pages={5064--5078},
  year={2022},
  publisher={IEEE}
}

@article{significance,
  title={A hitchhiker's guide to statistical comparisons of reinforcement learning algorithms},
  author={Colas, C{\'e}dric and Sigaud, Olivier and Oudeyer, Pierre-Yves},
  journal={arXiv preprint arXiv:1904.06979},
  year={2019}
}

@article{welch,
  title={The generalization of ‘STUDENT'S’problem when several different population varlances are involved},
  author={Welch, Bernard L},
  journal={Biometrika},
  volume={34},
  number={1-2},
  pages={28--35},
  year={1947},
  publisher={Oxford University Press}
}

@article{llmario,
  title={Policy Search through Genetic Programming and LLM-assisted Curriculum Learning},
  author={Jorgensen, Steven and Nadizar, Giorgia and Pietropolli, Gloria and Manzoni, Luca and Medvet, Eric and O'Reilly, Una-May and Hemberg, Erik},
  journal={ACM Transactions on Evolutionary Learning},
  year={2025},
  publisher={ACM New York, NY}
}

@Misc{QVACHER_EUROGP_2026,
  author = {Quentin Vacher and Nicolas Beuve  and Micka\"el Dardaillon and Karol Desnos},
  month  = April,
  title  = {EUROGP 2026 Artifacts},
  year   = {2026},
  url    = {https://github.com/gegelati/EUROGP-2026},
}

\end{document}